\RequirePackage{amsmath}         

\documentclass[runningheads,a4paper]{llncs}

\usepackage{graphicx}            
\usepackage{amssymb}             
\usepackage{amsfonts}            
\usepackage{enumitem}            
\usepackage{multirow}            
\usepackage{siunitx}             
\usepackage{url}                 
\usepackage{xspace}              
\usepackage[T1]{fontenc}         
\usepackage{hyperref}            
\usepackage{wrapfig}
\usepackage{todonotes}
\usepackage{units}
\usepackage{booktabs}

\graphicspath{{images/}}
\DeclareGraphicsExtensions{.pdf,.png,.jpg,.jpeg}

\newcommand{\seclabel}[1]{\label{sec:#1}}

\newcommand{\figlabel}[1]{\label{fig:#1}}

\newcommand{\eqnlabel}[1]{\label{eqn:#1}}

\newcommand{\figref}[1]{Fig.~\ref{fig:#1}\xspace}

\newcommand{\eqnref}[1]{(\ref{eqn:#1})\xspace}

\newcommand{\cmnew}{CM740\xspace}

\usepackage[firstpage=true]{background}
\newcommand\copyrighttext{%
	\parbox{\textwidth}{
		\centering
		\footnotesize
		 	In: RoboCup 2022: Robot World Cup XXV. LNCS 13561, Springer, May 2023.
	}
}

\SetBgContents{\copyrighttext}
\SetBgScale{1}
\SetBgColor{black}
\SetBgAngle{0}
\SetBgOpacity{1}
\SetBgPosition{current page.north}
\SetBgVshift{-2.5cm}
\SetBgHshift{3mm}

\begin{document}

\mainmatter

\title{RoboCup 2022 AdultSize Winner NimbRo: \\
	Upgraded Perception, Capture Steps Gait and Phase-based In-walk Kicks
}
\titlerunning{RoboCup 2022 AdultSize Winner NimbRo}

\author{Dmytro Pavlichenko, Grzegorz Ficht, Arash Amini, Mojtaba Hosseini, Raphael Memmesheimer, Angel Villar-Corrales, Stefan M. Schulz, Marcell Missura, Maren Bennewitz, and Sven Behnke}
\authorrunning{Pavlichenko, Ficht, Amini et al.}

\institute{Autonomous Intelligent Systems, Computer Science, Univ.\ of Bonn, Germany\\
\url{http://ais.uni-bonn.de}
\url{pavlichenko@ais.uni-bonn.de},
}

\maketitle

\begin{abstract}
Beating the human world champions by 2050 is an ambitious goal of the Humanoid League that provides a strong incentive for RoboCup teams to further improve and develop their systems. In this paper, we present upgrades of our system which enabled our team NimbRo to win the Soccer Tournament, the Drop-in Games, and the Technical Challenges in the Humanoid AdultSize League of RoboCup 2022. Strong performance in these competitions resulted in the Best Humanoid award in the Humanoid League. The mentioned upgrades include: hardware upgrade of the vision module, balanced walking with Capture Steps, and the introduction of phase-based in-walk kicks.
\end{abstract}

\section{Introduction}

\begin{figure}[!b]
	\centering
	\includegraphics[height=50mm]{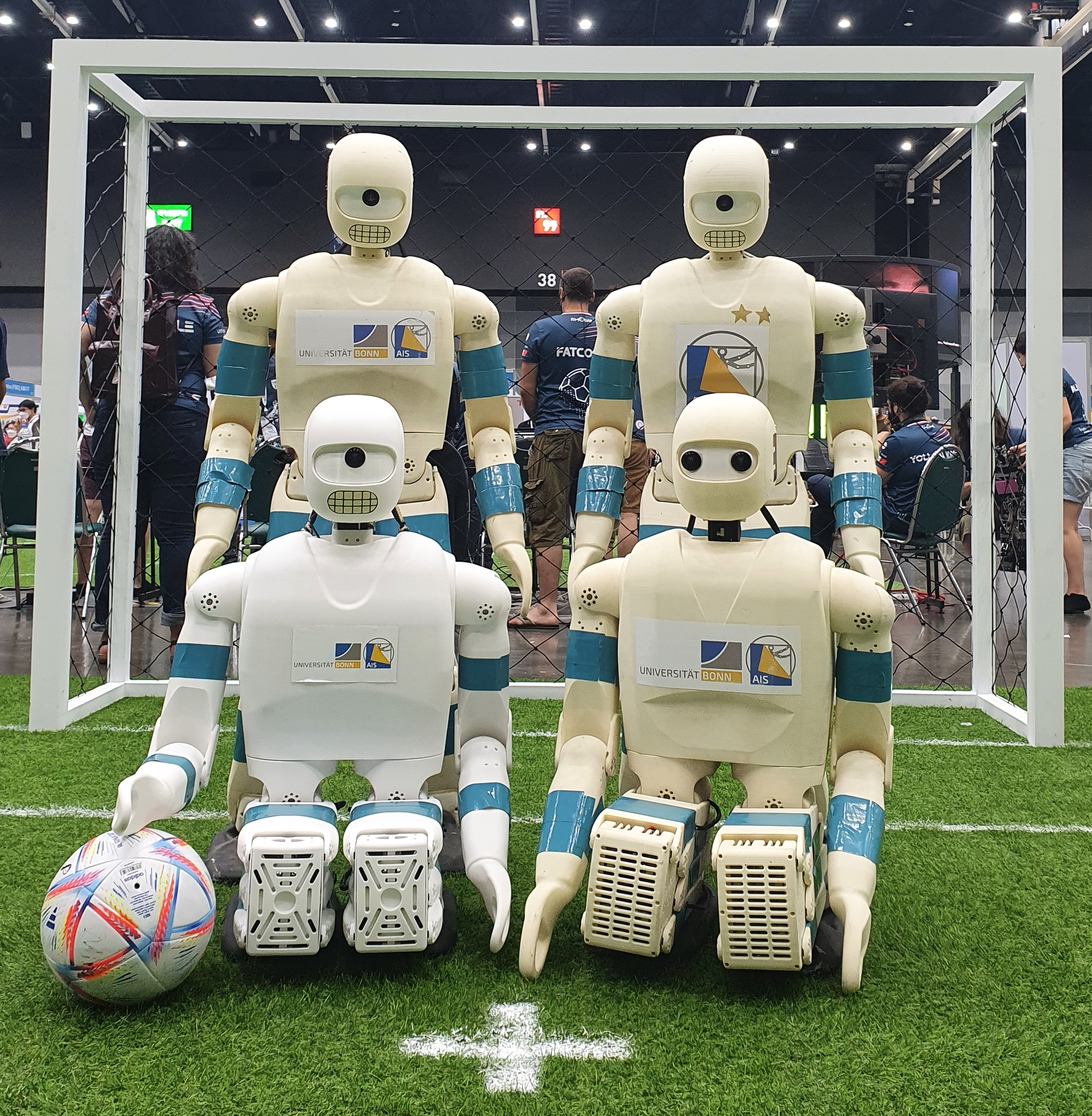}\hspace{0.005\linewidth}
	\includegraphics[height=50mm]{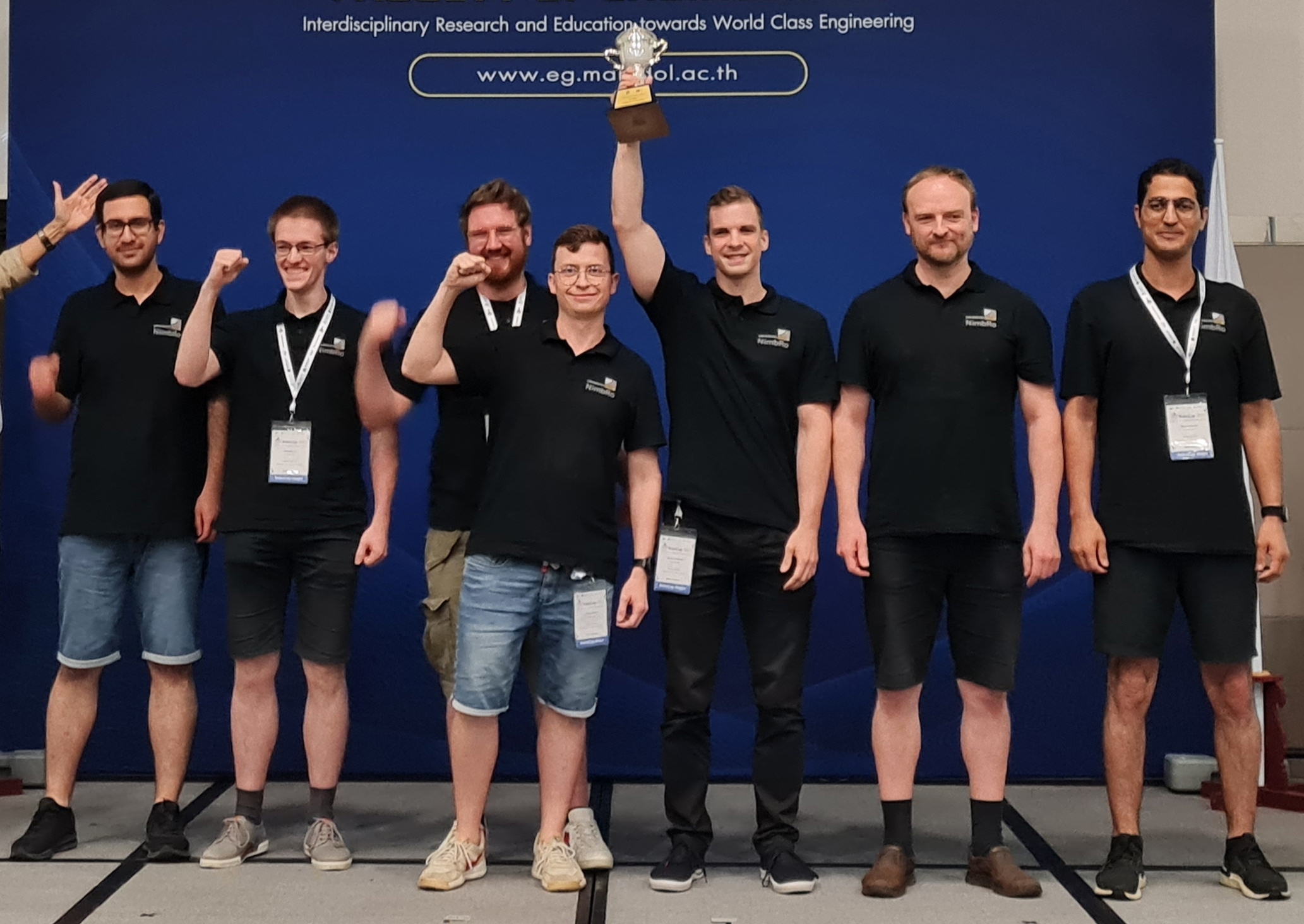}
	\caption{Left: NimbRo AdultSize Robots: NimbRo-OP2 and NimbRo-OP2X.
			Right: The NimbRo team at RoboCup 2022 in Bangkok, Thailand.}
	\figlabel{nimbro_team}
	\vspace{-4ex}
\end{figure}

The Humanoid AdultSize League works towards the vision of RoboCup: A team of robots winning against the human soccer world champion by 2050. Hence, the participating teams are challenged to improve their systems in order to reach this ambitious goal. Physical RoboCup championships have not been held in the previous two years, due to the COVID-19 pandemic. For RoboCup 2022, we upgraded our visual perception module with better cameras and more powerful GPUs as well as an improved deep neural network to enhance game perception and localization capabilities. We improved our gait based on the Capture Steps Framework to further enhance the stability of walking. Finally, introduction of phase-based in-walk kicks allowed for more ball control in close-range duels with opponents. We used three NimbRo-OP2X robots alongside with one NimbRo-OP2. Our robots came in first in all competitions of the Humanoid AdultSize League: the main Soccer Competition, Drop-in Games, and the Technical Challenges. Consequently, our robots won the Best Humanoid award in the Humanoid League. Our team is shown in \figref{nimbro_team}. The highlights of our performance during the RoboCup 2022 competition are available online\footnote{RoboCup 2022 NimbRo highlights video: \url{https://youtu.be/DfzkMawtSFA}}.

\section{NimbRo-OP2(X) Humanoid Robot Hardware}
\seclabel{robot_platforms}
For the competition in Bangkok, we have prepared four capable robots: one NimbRo-OP2~\cite{ficht2017nop2} 
and three NimbRo-OP2X robots. Both platforms are fully open-source in 
hardware\footnote{NimbRo-OP2X hardware: \url{https://github.com/NimbRo/nimbro-op2}} and software\footnote{NimbRo-OP2X software: \url{https://github.com/AIS-Bonn/humanoid_op_ros}},
with several publications~\cite{ficht2018nimbro,ficht2020nimbro} containing beneficial information on reproducing the platform. 
The robots, shown in \figref{nimbro_team} share similarities in naming, design and features, but are not identical. Both platforms feature a similar 
joint layout with 18 Degrees of Freedom~(DoF), with 5\,DoF per leg, 3\,DoF per arm, and 2\,DoF actuating the head. 
In contrast to other humanoid robots competing in the Humanoid League, the legs have a parallel kinematic structure 
that locks the pitch of the foot w.r.t. to the trunk~\cite{ficht2021bipedal}. The actuators receive commands 
from the ROS-based control framework through a \cmnew microcontroller board with a built-in six-axis IMU~(3-axis accelerometer \& 3-axis gyro).
As the hardware is based on off-the-shelf consumer-grade technology, the robots can be acquired at a fraction of the cost of 
similarly sized research-platforms~\cite{ficht2021bipedal} such as: ASIMO~\cite{shigemi2018asimo}, HRP-2~\cite{hrp2}, and HUBO~\cite{park2007mechanical}.

Apart from updating the OP2 actuator controller firmware to allow current-based torque measurements, 
its hardware was not changed from the RoboCup 2019 competition~\cite{nimbro_winners_2019}. The OP2X robots
were upgraded to accommodate for the weak points from the 2019 competition. After the field size increase,
perceiving objects from larger distances was challenging, as they would often be represented by single pixels. 
To address this issue, the 2MP Logitech C905 camera was replaced by a 5MP C930e camera. With an increase of the image capture resolution, the object perception improved greatly. The increase in resolution necessitated more computing power, for which we upgraded the GPU of each OP2X robot to the Nvidia RTX A2000. Applying these modifications was straightforward due to the modular design of the robot.

\section{Visual Perception of the Game Situation}
\seclabel{perception}

Our visual perception module benefits from both hardware and software improvements. First, upgrading the camera, equipped with a wide-angle lens, and GPU; second, supporting a dynamic resolution and enhancing our previous deep convolutional neural network to recognize soccer-related objects, including the soccer ball, goalposts, robots, field boundaries, and line segments.

\begin{figure}
    \centering
    \begin{tabular}{ccc}
        \includegraphics[width=0.32\linewidth]{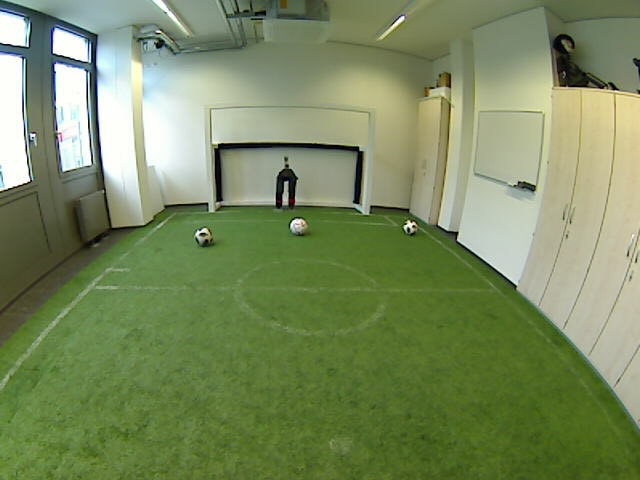} &
        \includegraphics[width=0.32\linewidth]{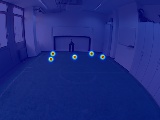} &
        \includegraphics[width=0.32\linewidth]{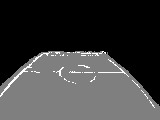} \\
        \includegraphics[width=0.32\linewidth]{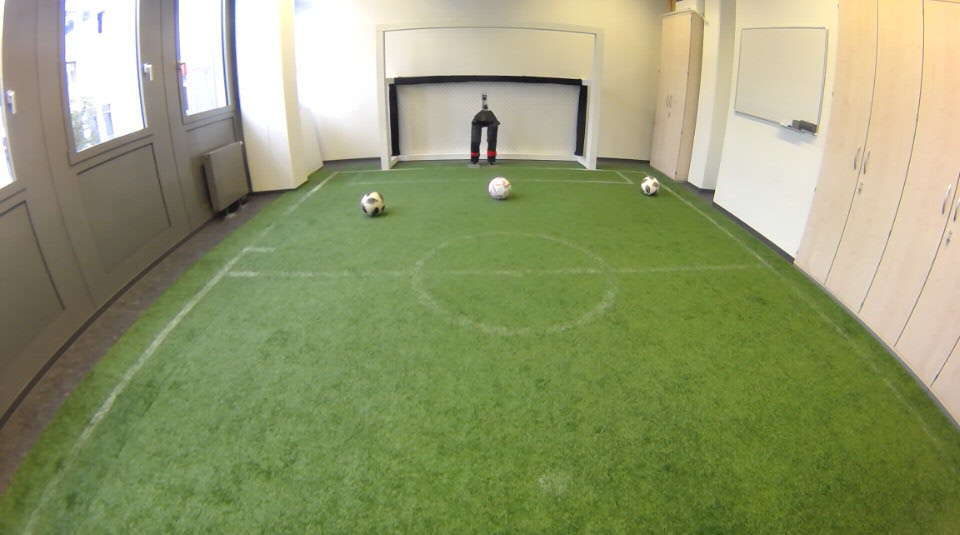} &
        \includegraphics[width=0.32\linewidth]{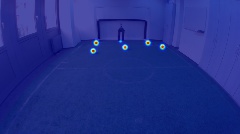} &
        \includegraphics[width=0.32\linewidth]{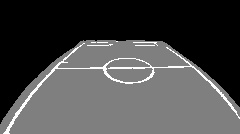} \\
    \end{tabular}
    \caption{An image example taken by Logitech C905 camera, heatmap, and segmentation predictions of NimbRoNet2 (first row). The same scene from the Logitech C930e camera, heatmap, and segmentation predictions of the new model (second row).}
    \label{fig:cameras}
\end{figure}

The upgraded GPU enabled us to use higher resolution images for the network input without decreasing the frame rate. Moreover, the new camera provided images with a high-quality and a wider horizontal field-of-view, compared to our previous Logitech C905 camera. We compare the two cameras by images captured from the same scene in~\figref{cameras}.

Inspired by our previous perception model (NimbRoNet2)~\cite{nimbro_winners_2019}, we used a deep convolutional neural network which is adapted from~\cite{amini2022humanoid}. Our model has an asymmetric U-Net architecture~\cite{unet2015} which consists of a pre-trained ResNet18 model~\cite{resnet2016} as an encoder and a decoder with a feature pyramid structure, where we only use the $1/4$ resolution, and three skip connections. This lighter model reduces inference time to 13\,ms by using bilinear upsampling in comparison to the 19\,ms runtime of NimbRoNet2. In addition, the new model yields better performance, as illustrated in~\figref{cameras}. The visual perception network has two prediction heads: object detection and segmentation. For object detection, we represent the object locations by heatmaps. We then apply a post-processing step to retrieve the locations from the estimated heatmaps.

We employ the mean squared error loss for training the heatmaps that represent the location of the balls, goalposts, and robots. Additionally, we use the cross entropy loss for training to predict the segmentation of background, field, and lines. The total loss is hence the linear combination of the two mentioned losses. As announced in RoboCup 2022, instead of using a single ball, 17 balls\footnote{\url{https://humanoid.robocup.org/hl-2022}} were selected as possible choices, of which three balls were used in the competition. To deal with this challenge, we exploited augmentation by randomly substituting the ball in the images with the announced balls.

Our pipeline shows the ability to cover the entire field and detects objects that are far away from the robot. Furthermore, since the localization module depends on landmarks (lines, center circle, and goalposts), the improvements of the vision perception module allow us to localize more precisely in the new field, where penalty area lines were added, through minor modifications in the localization module, e.g., increasing the weights of the center circle and goalposts.

\section{Robust Omnidirectional Gait with Diagonal Kick}

\subsection{Capture Step Walking}

\begin{figure}[t]
	\centering
	\includegraphics[width=0.13\linewidth]{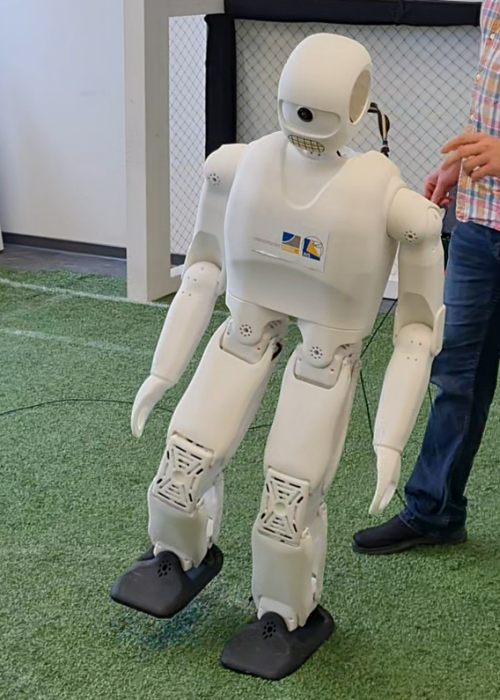}
	\includegraphics[width=0.13\linewidth]{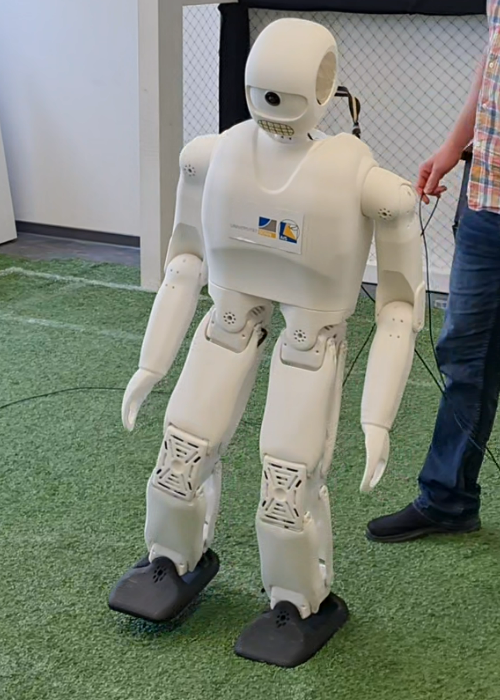}
	\includegraphics[width=0.13\linewidth]{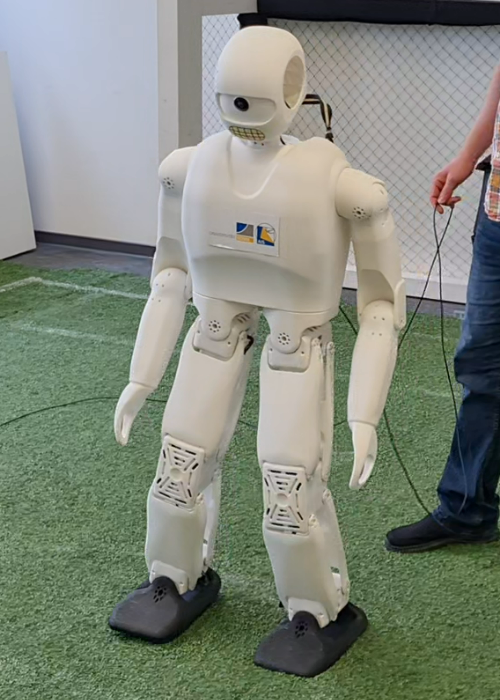}
	\includegraphics[width=0.13\linewidth]{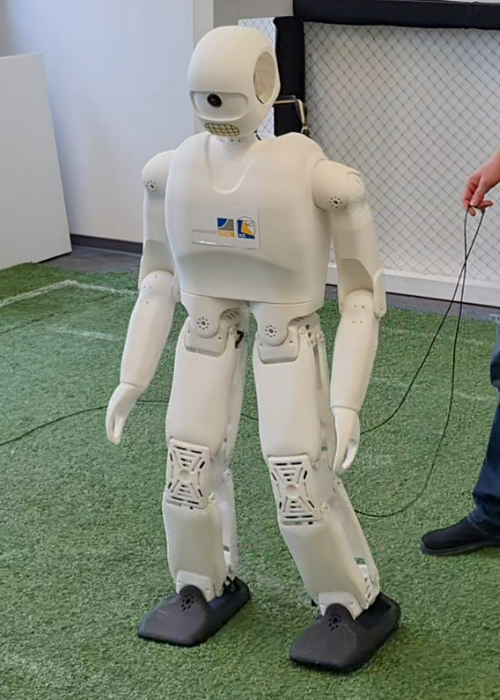}
	\includegraphics[width=0.13\linewidth]{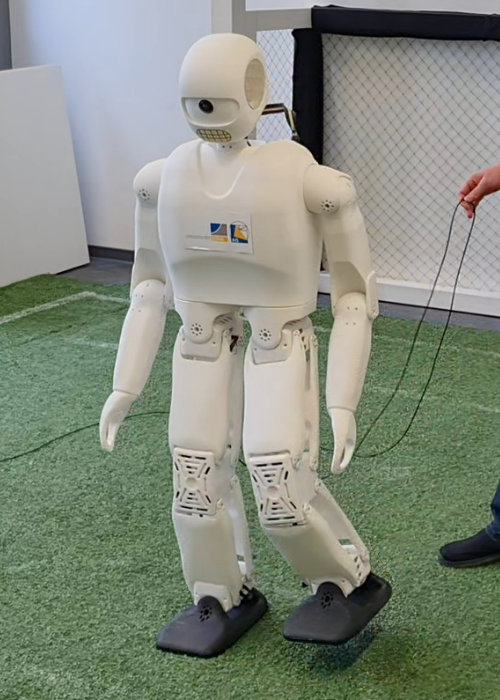}
	\includegraphics[width=0.13\linewidth]{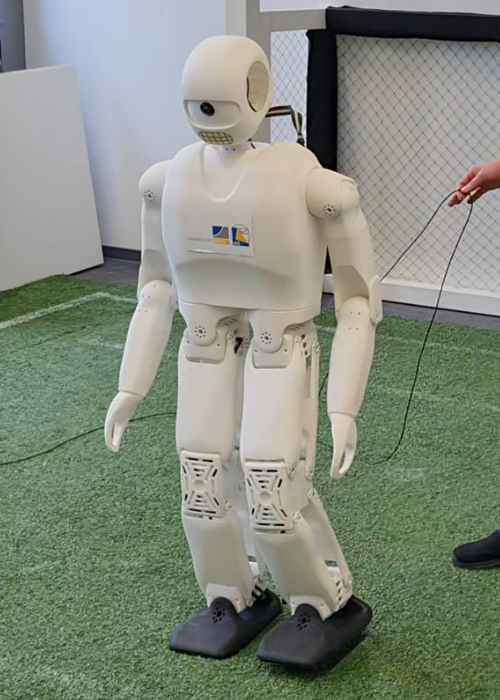}
	\includegraphics[width=0.13\linewidth]{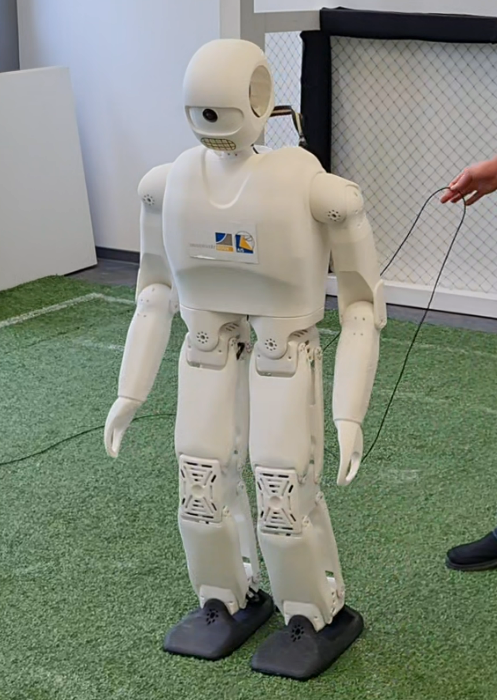}
	\includegraphics[height=70mm]{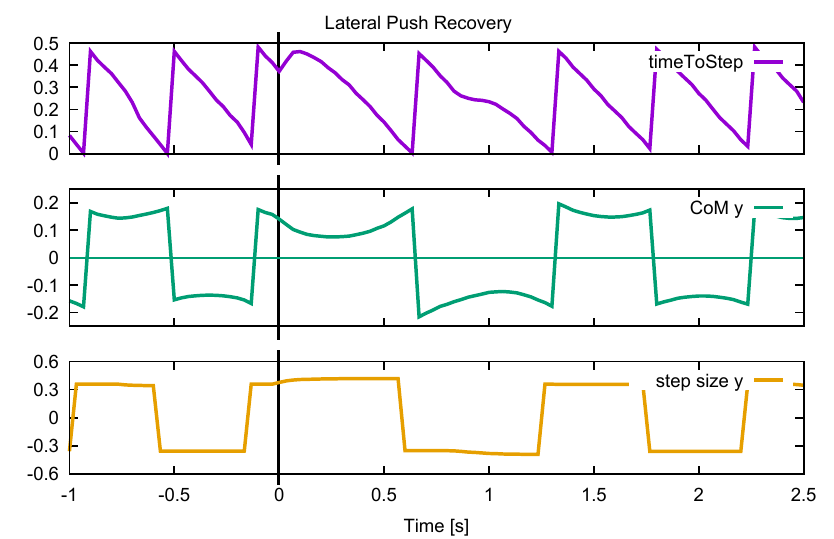}
	\caption{Lateral push recovery example. From top to bottom, the remaining time of a step, the lateral center of mass (CoM) location, and the lateral step size are plotted. The push occurs at time 0. The timing of the step is immediately adjusted for a longer step duration. The center of mass approaches the stance foot and then returns to the support exchange location at approx. 0.6\,s when the actual Capture Step is performed. The size of the Capture Step is adjusted from the default 30\,cm to 35\,cm. After the Capture Step at 1.25\,s, the gait returns to nominal values. }
	\figlabel{fig:lateral_recovery}
	\vspace{-4ex}
\end{figure}

\begin{figure}[t]
	\centering
	\includegraphics[width=0.13\linewidth]{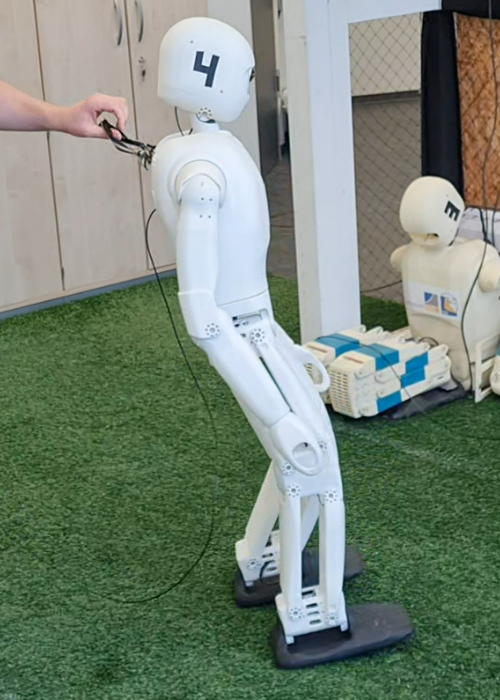}
	\includegraphics[width=0.13\linewidth]{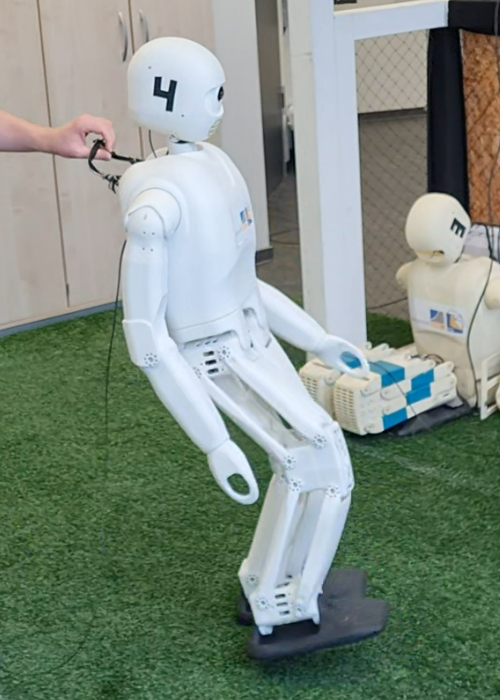}
	\includegraphics[width=0.13\linewidth]{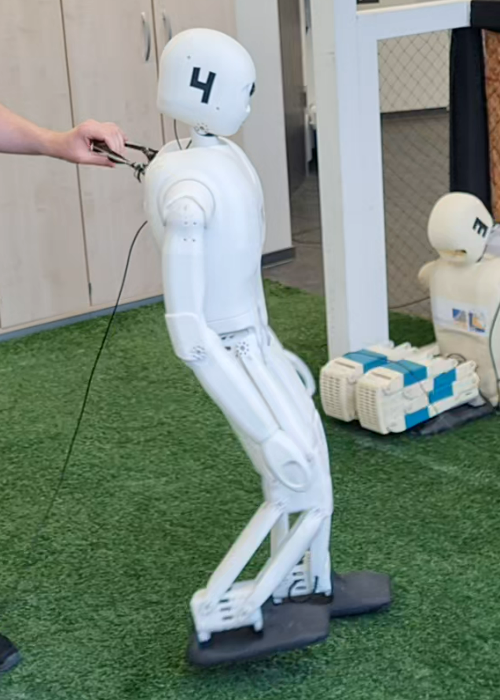}
	\includegraphics[width=0.13\linewidth]{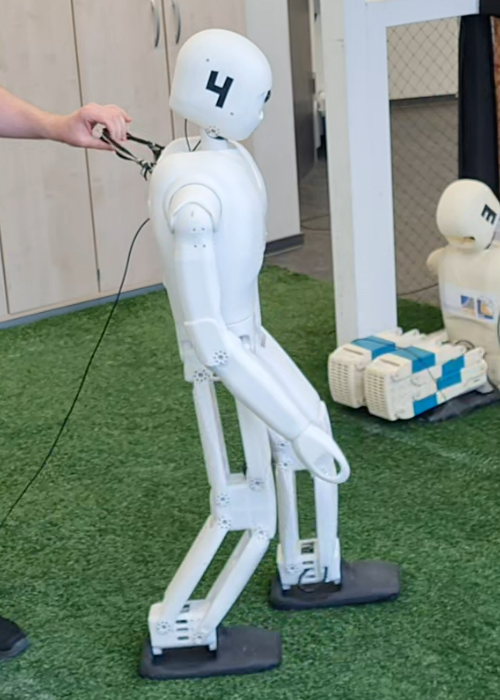}
	\includegraphics[width=0.13\linewidth]{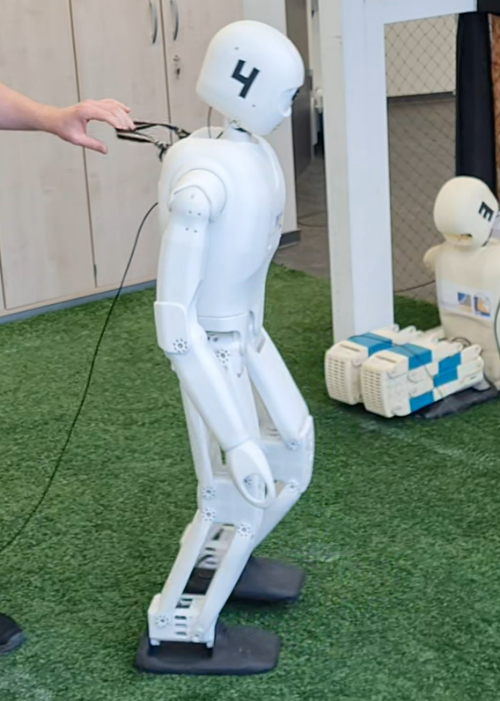}
	\includegraphics[width=0.13\linewidth]{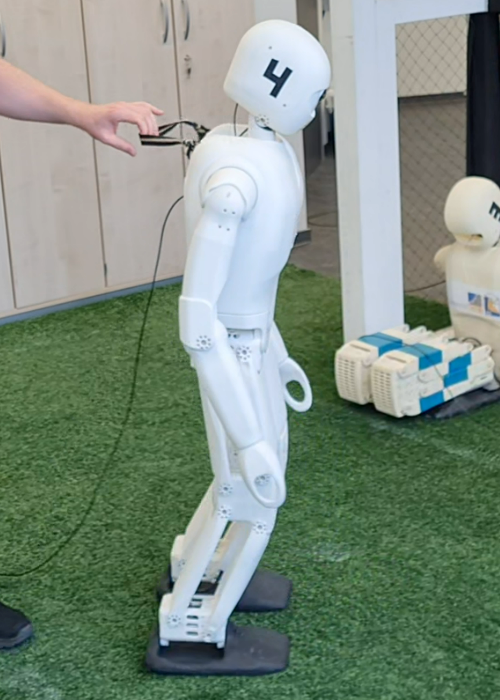}
	\includegraphics[width=0.13\linewidth]{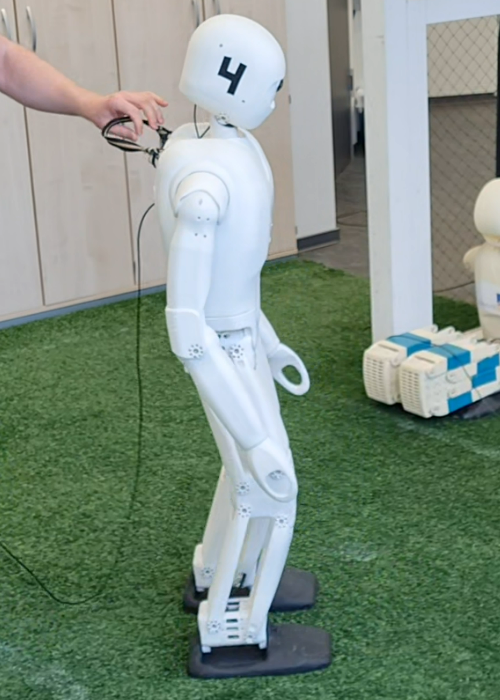}
	\includegraphics[height=70mm]{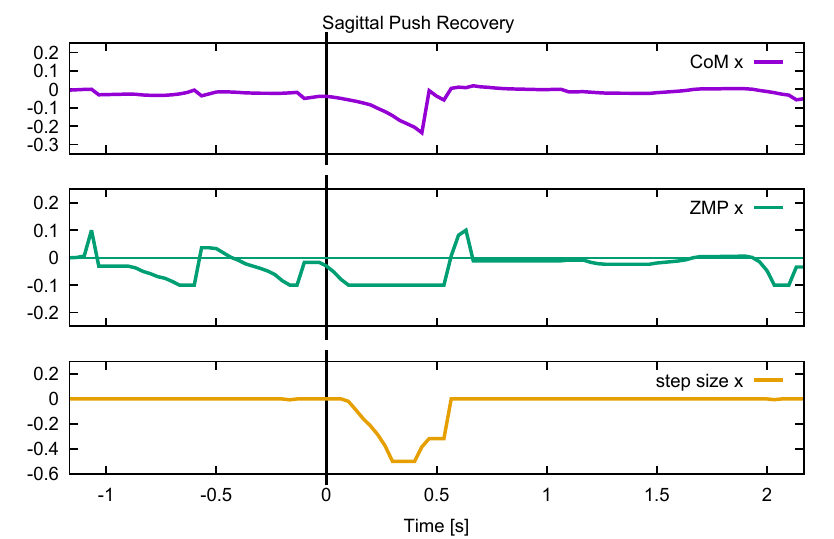}
	\caption{Sagittal push recovery example. From top to bottom, the sagittal center of mass (CoM) location, the sagittal zero moment point (ZMP) coordinate, and the sagittal step size are plotted. At the time 0 the robot experiences a backwards push. The ZMP immediately moves into the physical limit in the heel, and the step size is adjusted to perform a backwards step. After the first Capture Step is finished at 0.5\,s, the robot still has a small backwards momentum, which is nullified during the second recovery step and the gait returns to nominal values.}
	\figlabel{fig:sagittal_recovery}
	\vspace{-4ex}
\end{figure}

The walking of our robots is based on the Capture Step Framework \cite{Missura:CaptureSteps}. This framework combines an open loop gait pattern \cite{Missura:NimbRoGait}, that generates leg-swinging motions with a sinusoid swing pattern, with a balance control layer that models the robot as a linear inverted pendulum (LIP) and computes the timing, and location, of the next step in order to absorb disturbances and return the robot to an open-loop stable limit cycle. The parameters of the LIP are fitted to center of mass data recorded from the robot while the robot is walking open loop. The footstep locations computed by the LIP are then met by modifying the amplitude of the sinusoid leg swings during a step. The timing of the footsteps modulates the frequency of the central pattern generator such that the step motion is slowed down, or sped up, in order to touch the foot down at the commanded time. This year, for the first time, all of our robots were equipped with a Capture Step-capable walk. The Capture Steps proved especially useful for regaining balance after a collision with an opponent while fighting for the ball, for regaining balance after moving the ball with our seamlessly integrated in-walk kicks, and, of course, for winning the Push Recovery Challenge. \figref{fig:lateral_recovery} and \figref{fig:sagittal_recovery} show plots of a lateral and a sagittal push recovery example, respectively. 

\subsection{Balance State Estimation}

As with any real system, there is inherent noise in the IMU and joint sensors. Estimating balance-relevant state variables in presence of this noise
is a critical task, as the control is strictly tied to the state estimation. 

Before the 2022 competition~\cite{Missura:CaptureSteps}, the Center of Mass~(CoM) movement was estimated purely 
through a combination of joint encoders, kinematics, and a torso attitude estimator. 
The final output would be then smoothed with a Golay filter, also providing velocity and acceleration estimates.
This solution, while working to an extent, was not ideal. Noise would then be further suppressed in post-processing, 
along with balance-relevant dynamic effects, leading to the robot tending to walk open-loop at higher walking velocities.

We have adopted the idea of the Direct Centroidal Controller~(DCC) from~\cite{ficht2023direct} to use a Kalman Filter to estimate
the CoM state $\mathbf{c} = \begin{bmatrix}c &\dot{c} &\ddot{c}\end{bmatrix}$ on the sagittal $\mathbf{c}_x$ and lateral $\mathbf{c}_y$ planes
and supplementing the measurement model $\mathbf{z}_{k}$ with the unrotated and unbiased for gravity $g$ trunk acceleration~$^G\ddot{x}_t$ from the IMU:
\begin{equation} \eqnlabel{kalmanmeasurement}
\begin{bmatrix}^G\ddot{x}_t\\ ^G\ddot{y}_t\\ ^G\ddot{z}_t\end{bmatrix} = \begin{bmatrix}^T\ddot{x}_t\\ ^T\ddot{y}_t\\ ^T\ddot{z}_t\end{bmatrix}\mathbf{R_t}-\begin{bmatrix}0\\ 0\\ g\end{bmatrix},~
\mathbf{z}_{k,x} = \begin{bmatrix}1\quad0\\0\quad0\\0\quad1\end{bmatrix}\begin{bmatrix}c_x\\ ^G\ddot{x}_t \end{bmatrix} + \mathbf{v}_{k,x}.
\end{equation}
Unlike in the DCC, the mass position does not consider limb dynamics~\cite{ficht2020fast} and equates to a fixed point in the body frame.
This simplification is an advantage in the sensing scheme, as the accelerations of the trunk are directly linked to the trunk-fixed CoM. 
As a result, Capture Steps operated on the estimated state directly, allowing for precise and uniform balance control, independent of the walking speed.

\subsection{Phase-based In-walk Kick}

We introduced the "in-walk kick" approach in \cite{nimbro_winners_2019} and implemented it on NimbRo-OP2(X) platforms to compete at RoboCup 2019. 
The "in-walk kick" approach eliminated unnecessary stops to execute kicking motions, resulting in an improvement in the overall pace of the game.

\begin{figure}[!t]
	\centering
	\includegraphics[height=25mm]{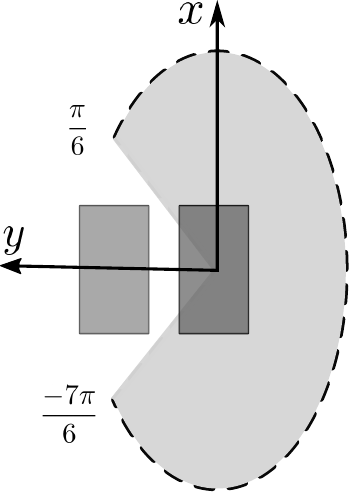}
	\caption{The feasible kick direction for the right leg. The dotted boundary represents the optimal ball position relative to the foot, shown as an ellipse from the configured optimal distance for the side and front kick. }
	\label{fig:kick-feasible-direction}
	\vspace{-4ex}
\end{figure}

However, the previously introduced approach had limitations that needed to be addressed. The strength of the kick depended on the accuracy of the foot placement behind the ball; the closer or farther the ball was from the optimal distance, the weaker the kicks were.
To shoot in the desired direction, the robot had to first rotate in that direction before executing the forward kick. For effective ball handling, the robot first had to align its foot behind the ball.

For RoboCup 2022, we addressed the above limitations to further improve the pace of the game.
Our approach allows the robot to perform kicks in feasible directions, while adjusting the foot behind the ball before the ball is actually kicked.
It also improves the strength of kicks where the ball is at a less than optimal distance from the foot.
The kick direction is not feasible if the leg is physically limited to move in that direction.
\figref{kick-feasible-direction} shows the feasible kick directions of the NimbRo-OP2 robot.

\begin{figure}[!t]
	\centering
	\includegraphics[height=38mm]{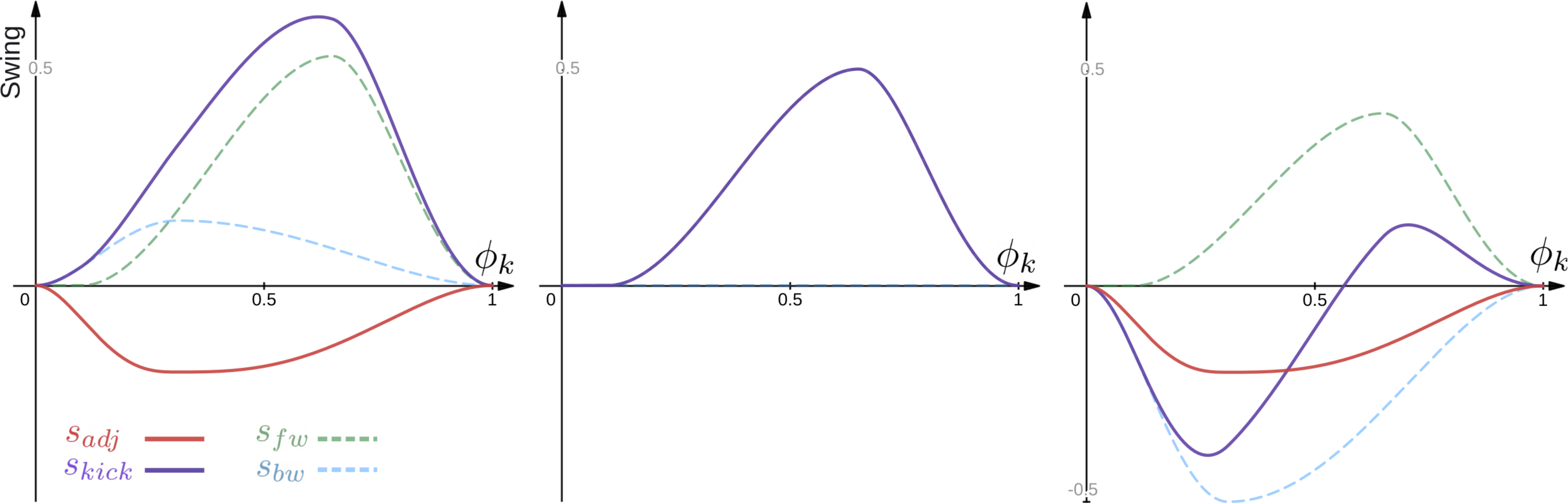}
	\caption{In-walk kick swing trajectories. $\phi_k$ is the kick phase (like $x_K$ in Fig. 6 of \cite{nimbro_winners_2019}). $s_{fw}$, $s_{bw}$, $s_{kick}$, and $s_{adj}$ are the forward, backward, kick, and adjust swings, respectively, formulated in \eqnref{kick-swings}. Three examples are given, from left to right: the ball farther away than the optimal distance with an offset along the $y$-axis, the ball at the optimal distance without $y$-offset, and the ball close to the foot with $y$-offset.}
	\label{fig:kick-traj}
	\vspace{-4ex}
\end{figure}

We define the kick coordinate frame such that the $x$ axis is along the direction of the kick and the origin of the frame is at the center of the kicking foot. The formulation of the kick trajectories is represented in this frame and later transformed into the local frame of the robot and applied to the sagittal and lateral trajectories of the leg, allowing the robot to perform diagonal and sidekicks in addition to the forward kick. We create two swing trajectories: the kick swing $s_{kick}$ and the adjust swing $s_{adj}$. The former applies a swing trajectory to the kick, the latter is a swing trajectory that adjusts the $y$ offset of the foot to the ball. \figref{kick-traj} shows three examples of the generated swing trajectories\footnote{Online in-walk kick graph: \url{https://www.desmos.com/calculator/v7wlvjtchl}}. The function of the swing trajectory $s$ is defined as:
\begin{equation}
\begin{split}
g\left(\phi,y_{0},y_{f},c\right) &=
y_{0}+\left(y_{f}-y_{0}\right)\left(6\left(1-\phi\right)^{2}\phi^{2}c+4\left(1-\phi\right)\phi^{3}+\phi^{4}\right),\\
s\left(\phi,\alpha,\phi_{p},c\right) &= 
\begin{cases}
g\left(\frac{\phi}{\phi_{p}},0,\alpha,c\right) & 0\le \phi < \phi_{p} \\
g\left(\frac{\phi-\phi_{p}}{1-\phi_{p}},\alpha,0,1-c\right) & \phi_{p}\le \phi\le 1
\end{cases},
\end{split}
\end{equation}
where $y_0$ and $y_f$ are the initial and final domains for the quartic Bezier curve $g$. The function $s$ for the phase variable $\phi$ represents a swing curve that reaches the peak amplitude $\alpha$ in the phase $\phi_{p}$ and returns to $0$ at the end of the phase while following a curvature defined by $c$.
The kick and swing trajectories are then formulated as follows:
\begin{equation}
\begin{split}
s_{fw}(\phi)   & = s\left(\phi,\alpha_{fw},\phi_{fw},c_{fw}\right), \\
s_{bw}(\phi)   & = s\left(\phi,\alpha_{bw},\phi_{bw},c_{bw}\right), \\
s_{kick}(\phi) & = s_{fw}(\phi) + s_{bw}(\phi), \\
s_{adj}(x)     & = s\left(\phi,\ \alpha_{y},\ \phi_{adj},\ c_{adj}\right),
\end{split} \label{eqn:kick-swings}
\end{equation}
where $s_{kick}$ denotes a swing trajectory for kicking the ball and $s_{adj}$ denotes an adjustment swing trajectory before kicking the ball.
$\alpha_{fw}=\alpha_{x}+0.8\left(\alpha_{opt}-\alpha_{x}\right)$ is the forward swing amplitude calculated with respect to the optimal swing amplitude $\alpha_{opt}$ and the ball position in the kick frame,
$\alpha_{bw}=\alpha_{fw}-\alpha_{opt}$ is the back-swing amplitude, and $c_{fw}$, $c_{bw}$, and $c_{adj}$ are the curvature gains of the forward, backward, and adjust swings, respectively.

With phase-based in-walk kicks, the behavior has more freedom to adjust the robot state before the kick and therefore makes faster decisions, which is especially beneficial in one-on-one fights where the opponent is also behind the ball and trying to kick it.

\section{Behavior Control}

For controlling the high-level behavior, which steers the robot behind the ball, avoids obstacles, and triggers inline kicks towards the opponent goal, we use a simple force field method. We are able to regard the robot as a holonomic point mass due to the omnidirectional control layer of our gait \cite{Missura:NimbRoGait} and can control it with a directional input vector whose direction determines the walking direction and its length determines the walking velocity. The vector has a separate component for turning, which is independent of the 2D walking direction on the plane. Using this interface, we can simply sum up forces that pull the robot towards the behind ball position and push the robot away from obstacles to obtain a suitable gait control vector. Orthogonal forces to the robot-ball line and to the robot-obstacle line help with circumventing the ball and walking around obstacles, respectively. The orientation of the robot is controlled by its own set of forces that rotate the robot first towards the ball, and then towards the ball target when the robot is near the behind ball position. Tuning parameters that determine the strength of the individual forces allow us to quickly adapt the behavior controller to different types of robots. This year, where all our robots were of a very similar build, we were able to use the same parameters for all robots. \figref{viewer} shows visualizations of components involved with the behavior control. There is no special goalie behavior since we find it more efficient to have two active field players.

\section{Debugging and Diagnostics}

\begin{figure}[!t]
	\centering
	\frame{\includegraphics[height=55mm]{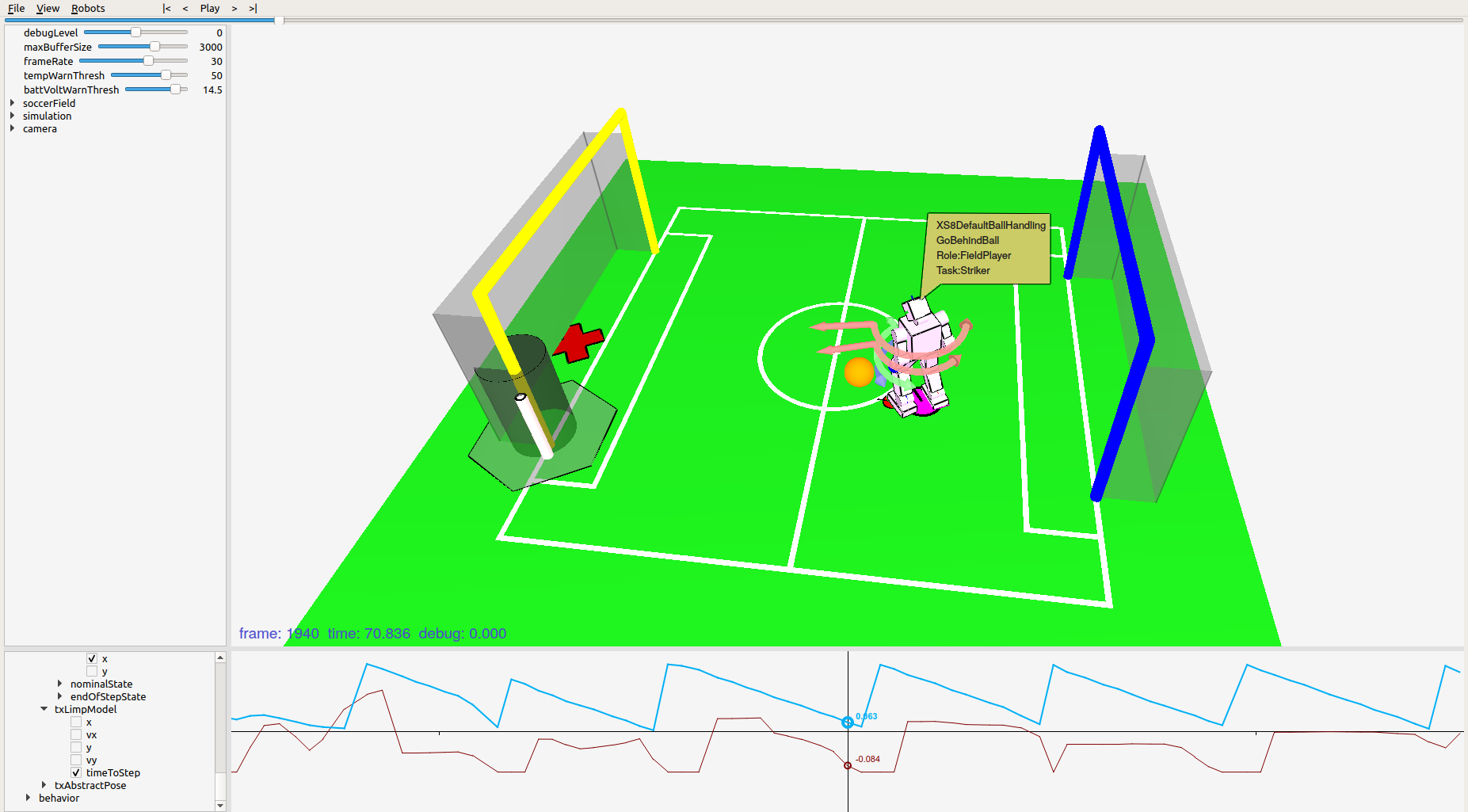}}\hspace{1.3mm}%
	\caption{The graphical debugging and diagnostics tool of team NimbRo. The 3D visualization area provides a large variety of information at one glimpse, such as: the whole-body state estimation of the robot, the localization, and the observed objects. The orange sphere is a ball. The red cross marks the spot towards which the robot should kick the ball. The small red circle is the behind ball position. The stack of arrows drawn around the robot model indicate the output of different control layers, where the top arrow is the force that controls the robot. The white vertical bar is the goalpost, and the gray cylinder around it represents an obstacle. The tool comes with an inbuilt plotter, and sliders for parameter tuning during operation.}
	\figlabel{viewer}
	\vspace{-4ex}
\end{figure}

As the complexity of the soccer-playing software increases over time, it is very important to develop visualization tools that help with debugging and allow for comprehensive diagnostics of hardware failures. \figref{viewer} shows a screenshot of our graphical analyzer. A 3D OpenGL scene shows the whole-body state estimation of our robot, the localization on the soccer field, observations such as the ball and the goal posts, the game state according to the game controller, the role and task of the robot, and information about the behavior and gait controller output. Obstacles are marked with a black cylinder. Our tool also includes a plotter for variables that cannot be easily visualized, such as the estimated time until the next support exchange shown by the blue curve in the plot area at the bottom of the screen. The communication interface between the debugging tool and the robot seamlessly integrates into our ROS infrastructure.

\section{Technical Challenges}
\seclabel{technical_challenges}

In this section, we describe our approach to Technical Challenges, which is a separate competition within AdultSize League. In a 25 minutes long time slot, the robots are required to perform four individual tasks: Push Recovery, Parkour, High Kick, and Goal Kick from Moving Ball. The strict time restriction strongly advocates for robust and reliable solutions.

\subsection{Push Recovery}
\seclabel{push_recovery}

In this challenge, the robot has to withstand three pushes in a row from the front and from the back while walking on the spot. The pushes are induced by a pendulum which hits the robot at the height of the CoM. The robots are ranked by the combination of pendulum weight and pendulum retraction distance, normalized by the robot mass. Our robot managed to withstand pushes from both 3\,kg and 5\,kg pendulums, thanks to the Capture Steps Framework, winning in this challenge.

\subsection{Parkour}
\seclabel{parkour}

\begin{figure}
	\centering
	\includegraphics[width=0.16\linewidth]{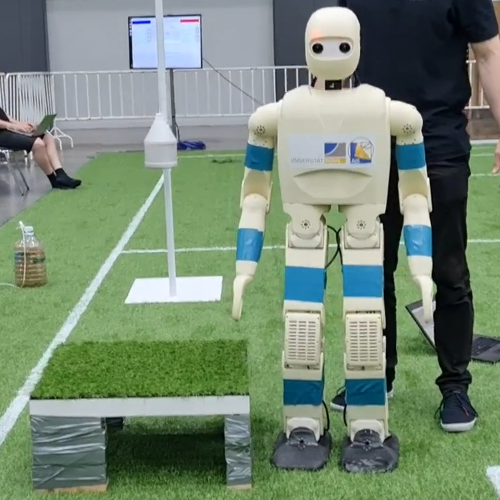}
	\includegraphics[width=0.16\linewidth]{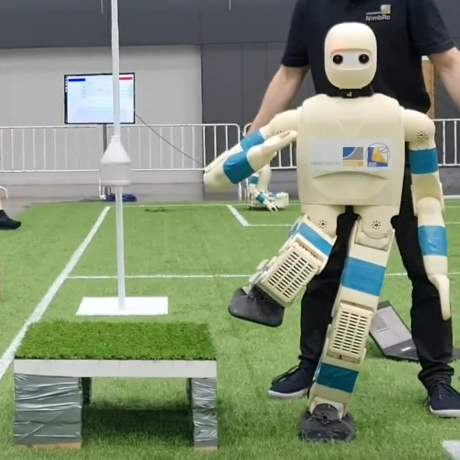}
	\includegraphics[width=0.16\linewidth]{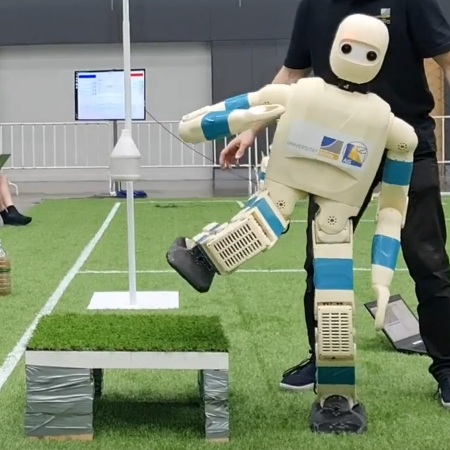}
	\includegraphics[width=0.16\linewidth]{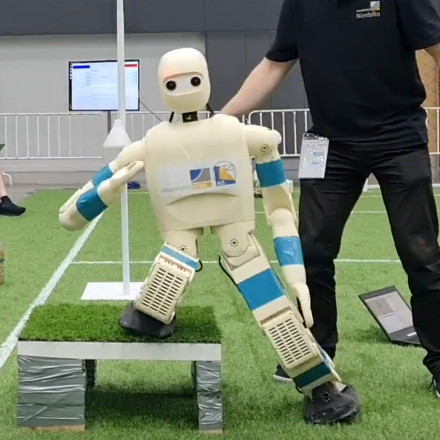}
	\includegraphics[width=0.16\linewidth]{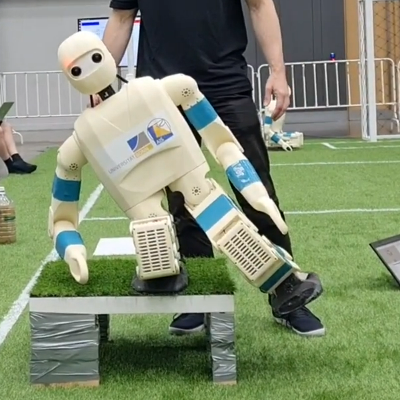}
	\includegraphics[width=0.16\linewidth]{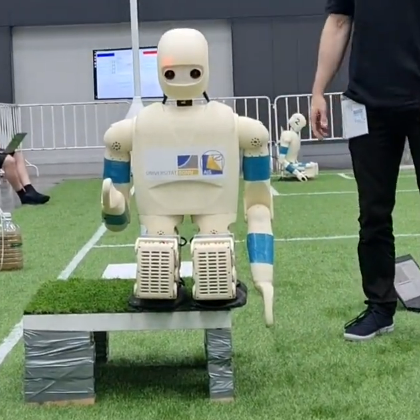}
	\caption{Technical Challenge: Parkour.}
	\label{fig:parkour}
	\vspace*{-4ex}
\end{figure}

In this challenge, the robot has to go up a platform and then go back down. The robots are ranked by the height of the platform. We performed this challenge using a manually designed motion. Since our robots have parallel leg kinematics, it was advantageous for us to go on top of the platform with a side-step, where our robots have more mobility. The motion included a controlled fall on the platform, allowing the foot to land closer to the center of the platform, creating space for the other foot in the later stage of the ascent. During this phase, the gains of the actuators were reduced, providing a damping effect when landing on the platform. Then, a series of CoM shifts brought the other foot on the platform, achieving a stable stance (\figref{parkour}). We did not descend from the platform, because such motion imposed a significant load on the motors. Our robot managed to go up a 30\,cm high platform, coming in second in this challenge.

\subsection{High Kick}
\seclabel{high_kick}

In this challenge, the goal is to score a goal by kicking the ball over an obstacle. The teams are ranked by the height of the obstacle in a successful attempt. The ball starts at the penalty mark and the obstacle is at the goal line. In order to reliably kick over the obstacle, we first perform a kick of low strength to move the ball closer to the obstacle. Then, we kick over it utilizing a kick motion with the foot making contact with the ball as low as possible and using a "scoop"-shaped foot. Our robot came in first, kicking over a 16\,cm high obstacle.

\subsection{Goal Kick from Moving Ball}
\seclabel{moving_ball}

\begin{figure}
	\centering
	\includegraphics[width=0.24\linewidth]{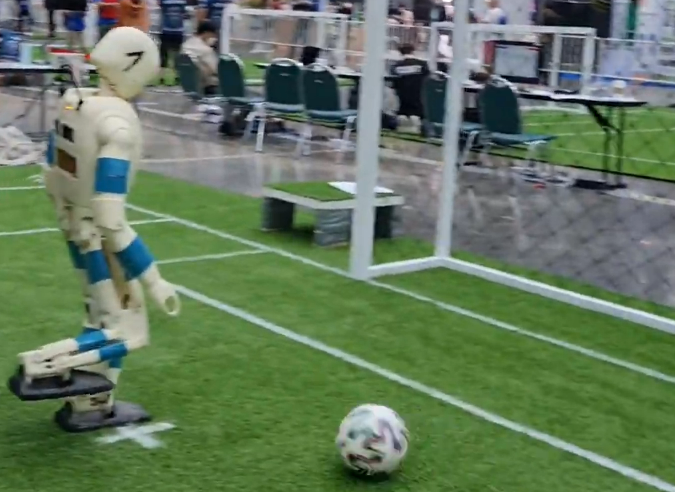}
	\includegraphics[width=0.24\linewidth]{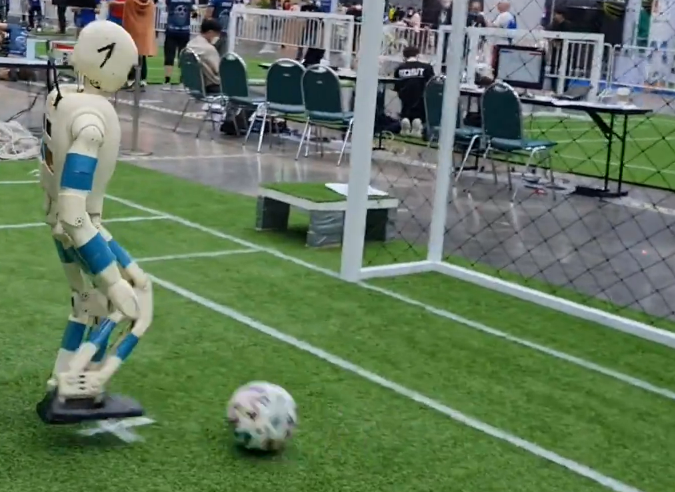}
	\includegraphics[width=0.24\linewidth]{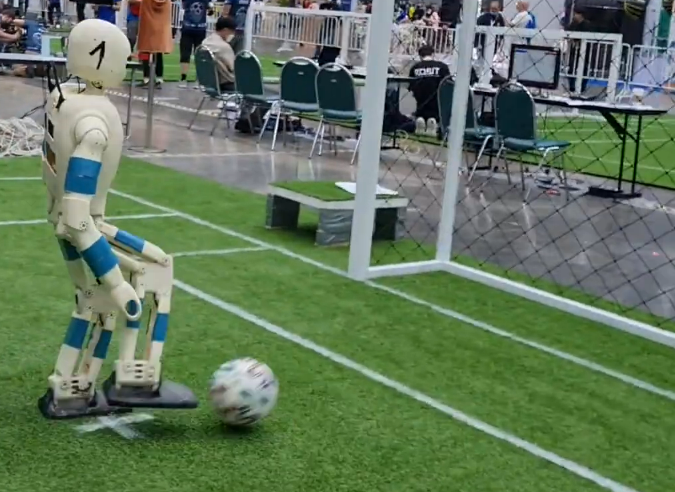}
	\includegraphics[width=0.24\linewidth]{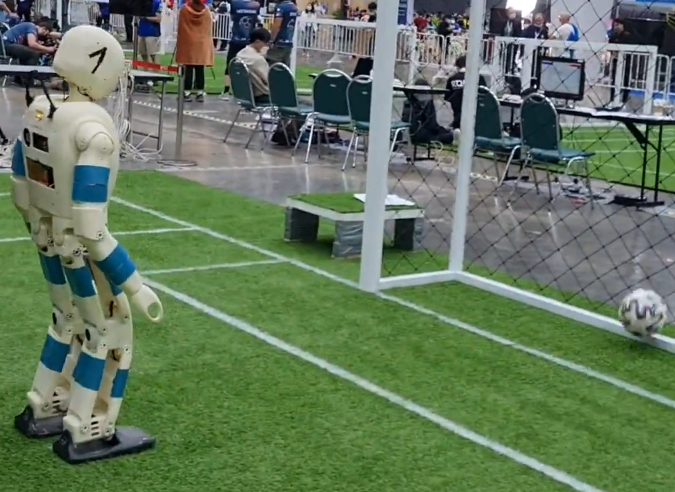}
	\caption{Technical Challenge: Goal Kick from Moving Ball.}
	\label{fig:moving_ball}
	\vspace*{-4ex}
\end{figure}

In this challenge, a robot has to score a goal by kicking a moving ball. The robot is stationary at the penalty mark and the ball is at the corner. The ball is passed to the robot by a human. The teams are ranked by the number of goals scored in three successive attempts. In order to reliably score goals from a moving ball, we estimate the ball arrival time from velocity and acceleration estimates, which are calculated from the series of ball detections. This enables our robots to start kicking at the right moment, when the ball is approaching the foot (\figref{moving_ball}). Our robot came in first in this challenge, scoring in three successive attempts.

\section{Soccer Game Performance}

At the RoboCup 2022 AdultSize soccer competition, robots played 2 vs. 2 soccer games autonomously on a $9 \times 14$\,m soccer field. In addition to the main tournament, there were Drop-in games, where robots from different teams formed temporary joint teams to compete in 2 vs. 2 games. Our robots performed well and won the AdultSize tournament, which included six round-robin games and the finals, with a total score of 43:2, winning the final 7:1. Our robots also won the Drop-In competition, accumulating 22 points compared to 5.5 points of the second-best team, finishing three games with a total score of 15:3. During the competition, our robots played 10 games with a total score of 58:5 and total duration of 200\,min.

\section{Conclusions}

In this paper, we presented improvements of hardware and software components of our humanoid soccer robots which led us to winning all available competitions in the AdultSize league of RoboCup 2022 in Bangkok: main tournament, Drop-In games, and Technical Challenges, where our robots scored in each challenge. Consistently strong performance of our robots through the competition resulted in receiving the Best Humanoid Award. Improved components include: upgraded perception module, Capture Steps gait, and phase-based in-walk kicks. These innovations allowed for improved localization and ball perception, more robust walking, and more dynamic ball handling --- contributing to winning the RoboCup 2022 AdultSize Humanoid League.

\subsection*{Acknowledgments}
\footnotesize
This work was partially funded by H2020 project EUROBENCH, GA 779963.

\bibliographystyle{splncs03}
\bibliography{winners_2022}

\begin{thebibliography}{10}
\providecommand{\url}[1]{\texttt{#1}}
\providecommand{\urlprefix}{URL }
\providecommand{\doi}[1]{https://doi.org/#1}

\bibitem{amini2022humanoid}
Amini, A., Farazi, H., Behnke, S.: Real-time pose estimation from images for
  multiple humanoid robots. In: RoboCup 2021: Robot World Cup XXIV. pp.
  91--102. Springer LNCS 13132 (2022)

\bibitem{ficht2017nop2}
Ficht, G., Allgeuer, P., Farazi, H., Behnke, S.: {N}imb{R}o-{O}{P}2: {G}rown-up
  3{D} printed open humanoid platform for research. In: 17th IEEE-RAS Int.
  Conf. on Humanoid Robots (Humanoids) (2017)

\bibitem{ficht2020fast}
Ficht, G., Behnke, S.: Fast whole-body motion control of humanoid robots with
  inertia constraints. In: IEEE Int. Conf. on Robotics and Automation (ICRA).
  pp. 6597--6603 (2020)

\bibitem{ficht2021bipedal}
Ficht, G., Behnke, S.: Bipedal humanoid hardware design: {A} technology review.
  Current Robotics Reports  \textbf{2}(2),  201--210 (2021)

\bibitem{ficht2023direct}
Ficht, G., Behnke, S.: Direct centroidal control for balanced humanoid
  locomotion. In: 25th International Conference on Climbing and Walking Robots
  (CLAWAR). pp. 242--255 (2022)

\bibitem{ficht2018nimbro}
Ficht, G., Farazi, H., Brandenburger, A., Rodriguez, D., Pavlichenko, D.,
  Allgeuer, P., Hosseini, M., Behnke, S.: Nimb{R}o-{O}{P}2{X}: {A}dult-sized
  open-source {3D} printed humanoid robot. In: 18th IEEE-RAS Int. Conf. on
  Humanoid Robots (Humanoids) (2018)

\bibitem{ficht2020nimbro}
Ficht, G., Farazi, H., Rodriguez, D., Pavlichenko, D., Allgeuer, P.,
  Brandenburger, A., Behnke, S.: {Nimbro-OP2X:} {A}ffordable adult-sized
  {3D}-printed open-source humanoid robot for research. Int. J. of Humanoid
  Robotics  \textbf{17}(05) (2020)

\bibitem{resnet2016}
He, K., Zhang, X., Ren, S., Sun, J.: Deep residual learning for image
  recognition. In: IEEE Int. Conf. on Computer Vision and Pattern Recognition
  (CVPR). pp. 770--778 (2016)

\bibitem{hrp2}
Hirukawa, H., Kanehiro, F., Kaneko, K., Kajita, S., Fujiwara, K., Kawai, Y.,
  Tomita, F., Hirai, S., Tanie, K., Isozumi, T., et~al.: Humanoid robotics
  platforms developed in {HRP}. Robotics and Autonomous Systems pp. 165--175
  (2004)

\bibitem{Missura:NimbRoGait}
Missura, M., Behnke, S.: Self-stable omnidirectional walking with compliant
  joints. In: Workshop on Humanoid Soccer Robots, IEEE-RAS International
  Conference on Humanoid Robots (Humanoids). Atlanta, USA (2013)

\bibitem{Missura:CaptureSteps}
Missura, M., Bennewitz, M., Behnke, S.: Capture steps: Robust walking for
  humanoid robots. Int. J. Humanoid Robotics  \textbf{16}(6),
  1950032:1--1950032:28 (2019)

\bibitem{park2007mechanical}
Park, I.W., Kim, J.Y., Lee, J., Oh, J.H.: Mechanical design of the humanoid
  robot platform, hubo. Advanced Robotics  \textbf{21}(11),  1305--1322 (2007)

\bibitem{nimbro_winners_2019}
Rodriguez, D., Farazi, H., Ficht, G., Pavlichenko, D., Brandenburger, A.,
  Hosseini, M., Kosenko, O., Schreiber, M., Missura, M., Behnke, S.: {RoboCup}
  2019 {AdultSize} winner {NimbRo}: {Deep} learning perception, in-walk kick,
  push recovery, and team play capabilities. In: RoboCup 2019: Robot World Cup
  XXIII. pp. 631--645. Springer LNCS 11531 (2019)

\bibitem{unet2015}
Ronneberger, O., Fischer, P., Brox, T.: U-{N}et: {C}onvolutional networks for
  biomedical image segmentation. In: Medical Image Computing and
  Computer-Assisted Intervention (MICCAI). pp. 234--241. Springer (2015)

\bibitem{shigemi2018asimo}
Shigemi, S., Goswami, A., Vadakkepat, P.: Asimo and humanoid robot research at
  honda. Humanoid robotics: A reference pp. 55--90 (2018)

\end{thebibliography}

\end{document}